\documentclass[10pt,twocolumn,letterpaper]{article}

\usepackage{wacv}
\usepackage{times}
\usepackage{epsfig}
\usepackage{graphicx}
\usepackage{amsmath}
\usepackage{amssymb}
\usepackage{multirow}
\usepackage{balance}
\usepackage{booktabs}
\usepackage{color}
\usepackage{array}
\usepackage{tabu}
\usepackage{comment}

\wacvfinalcopy
\def\wacvPaperID{143} 
\graphicspath{{./}{./figure/}}

\def\Vec#1{{\boldsymbol{#1}}}

\begin{document}
\title{Matching Image Sets via Adaptive Multi Convex Hull}

\author
  {
  {\it Shaokang Chen, Arnold Wiliem, Conrad Sanderson, Brian C. Lovell}\\
  ~\\
  University of Queensland, School of ITEE, QLD 4072, Australia\\
  NICTA, GPO Box 2434, Brisbane, QLD 4001, Australia\\
  Queensland University of Technology, Brisbane, QLD 4000, Australia
  }

\maketitle
\thispagestyle{empty}
\pagestyle{empty}

\begin{abstract}

\vspace{-2ex}
\noindent
Traditional nearest points methods use all the samples in an image
set to construct a single convex or affine hull model for
classification. However, strong artificial features and noisy data
may be generated from combinations of training samples when
significant intra-class variations and/or noise occur in the image
set. Existing multi-model approaches extract local models by
clustering each image set individually only once, with fixed
clusters used for matching with various image sets. This may not
be optimal for discrimination, as undesirable environmental
conditions (eg.~illumination and pose variations) may
result in the two closest clusters representing different
characteristics of an object (eg.~frontal face being compared to non-frontal face).
To address the above problem, we propose a
novel approach to enhance nearest points based methods by
integrating affine/convex hull classification with an adapted
multi-model approach. We first extract multiple local convex hulls
from a query image set via maximum margin clustering to diminish
the artificial variations and constrain the noise in
local convex hulls. We then propose adaptive reference clustering
(ARC) to constrain the clustering of each gallery image set by
forcing the clusters to have resemblance to the clusters in the
query image set. By applying ARC, noisy clusters in the query 
set can be discarded. Experiments on Honda, MoBo and
\mbox{ETH-80} datasets show that the proposed method outperforms
single model approaches and other recent techniques, such
as Sparse Approximated Nearest Points,
Mutual Subspace Method and Manifold Discriminant Analysis.
\end{abstract}

\vspace{-3ex}
\section{Introduction}
\vspace{-1ex}
Compared to single image matching techniques, image set matching approaches
exploit set information for improving discrimination accuracy as well as robustness to image
variations, such as pose, illumination and misalignment~\cite{Arandjelovic05,Hakan10,Tae-Kyun07,Ruiping09}.
Image set classification techniques can be categorised into two general classes:
parametric and non-parametric methods.
The former represent image sets with parametric distributions~\cite{Arandjelovic05,Cardinaux_TSP_2006,Honda-dataset}.
The distance between two sets can be measured
by the similarity between the estimated parameters of the distributions.
However, the estimated parameters might be dissimilar if the
training and test data sets of the same subject have weak statistical correlations~\cite{Tae-Kyun07,MMD08}.

State-of-the-art non-parametric methods can be categorised into two groups:
single-model and multi-model methods.
Single-model methods attempt to represent sets as linear subspaces~\cite{Tae-Kyun07,MSM-98},
or affine/convex hulls~\cite{Hakan10,SANP11}.
For single linear subspace methods, principal angles are generally used to measure the difference between two subspaces~\cite{Sanderson_AVSS_2012,MSM-98}.
As the similarity of data structures is used for comparing sets,
the subspace approach can be robust to noise and relatively small number of samples~\cite{MMD08,MSM-98}.
However, single linear subspace methods consider the structure of all data samples without selecting optimal subsets for classification.

\begin{figure}[!b]
  \vspace{-2ex}
  \centering
  \begin{minipage}{1.0\columnwidth}

    \begin{minipage}{0.05\textwidth}
      \begin{minipage}{\textwidth}
        \vspace{1ex}
      \end{minipage}
      \begin{minipage}{\textwidth}
        \footnotesize
        \mbox{\bf (a)}
        \vspace{6ex}
      \end{minipage}
      \begin{minipage}{\textwidth}
        \footnotesize
        \vspace{6ex}
        \mbox{\bf (b)}
      \end{minipage}
      \begin{minipage}{\textwidth}
        \vspace{1ex}
      \end{minipage}
    \end{minipage}
    \hfill
    \begin{minipage}{0.17\textwidth}
      \begin{minipage}{\textwidth}
      \centering
      $I_1$
      \end{minipage}
      \begin{minipage}{\textwidth}
        \centering
        \includegraphics[width=1\textwidth]{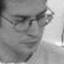}
      \end{minipage}
      \begin{minipage}{\textwidth}
        \centering
        \includegraphics[width=1\textwidth]{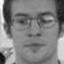}
      \end{minipage}
      \begin{minipage}{\textwidth}
        \vspace{1ex}
      \end{minipage}
    \end{minipage}
    \vrule
    \hspace{0.125ex}
    \begin{minipage}{0.17\textwidth}
      \begin{minipage}{\textwidth}
      \centering
      $I_{gen}$
      \end{minipage}
      \begin{minipage}{\textwidth}
        \centering
        \includegraphics[width=1\textwidth]{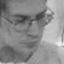}
      \end{minipage}
      \begin{minipage}{\textwidth}
        \centering
        \includegraphics[width=1\textwidth]{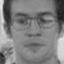}
      \end{minipage}
      \begin{minipage}{\textwidth}
        $w = 0.25$
      \end{minipage}
    \end{minipage}
    \hfill
    \begin{minipage}{0.17\textwidth}
      \begin{minipage}{\textwidth}
      \centering
      $I_{gen}$
      \end{minipage}
      \begin{minipage}{\textwidth}
        \centering
        \includegraphics[width=1\textwidth]{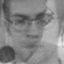}
      \end{minipage}
      \begin{minipage}{\textwidth}
        \centering
        \includegraphics[width=1\textwidth]{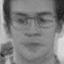}
      \end{minipage}
      \begin{minipage}{\textwidth}
        $w = 0.5$
      \end{minipage}
    \end{minipage}
    \hfill
    \begin{minipage}{0.17\textwidth}
      \begin{minipage}{\textwidth}
      \centering
      $I_{gen}$
      \end{minipage}
      \begin{minipage}{\textwidth}
        \centering
        \includegraphics[width=1\textwidth]{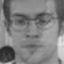}
      \end{minipage}
      \begin{minipage}{\textwidth}
        \centering
        \includegraphics[width=1\textwidth]{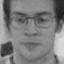}
      \end{minipage}
      \begin{minipage}{\textwidth}
        $w = 0.75$
      \end{minipage}
    \end{minipage}
    \vrule
    \hspace{0.125ex}
    \begin{minipage}{0.17\textwidth}
      \begin{minipage}{\textwidth}
      \centering
      $I_2$
      \end{minipage}
      \begin{minipage}{\textwidth}
        \centering
        \includegraphics[width=1\textwidth]{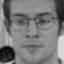}
      \end{minipage}
      \begin{minipage}{\textwidth}
        \centering
        \includegraphics[width=1\textwidth]{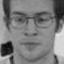}
      \end{minipage}
      \begin{minipage}{\textwidth}
        \vspace{1ex}
      \end{minipage}
    \end{minipage}
  \end{minipage}
  \vspace{0.5ex}
  \caption
    {
    Illustration of artificially generated images from two training samples $I_1$ and $I_2$ via $I_{gen} = (1-w) I_{1} + w I_{2}$.
    For images in a row, the three images (denoted as $I_{gen}$) in the middle are generated from convex combinations of the first (denoted as $I_{1}$) and the last (denoted as $I_{2}$) images.
    Case {\bf (a)}: as $I_{1}$ and $I_{2}$ are very different from each other, the generated images may contain unrealistic artificial features, such as the profile in the middle of the face in the third and fourth image.
    Case {\bf (b)}: as $I_{1}$ and $I_{2}$ are similar to each other, the three generated images contain only minor artificial features.
    }
  \label{Fig:linear-generation}
\end{figure}

Convex hull approaches use geometric distances (eg.~Euclidean
distance between closest points) to compare sets. Given two sets,
the closest points between two convex hulls are calculated by
least squares optimisation. As such, these methods adaptively
choose optimal samples to obtain the distance between sets,
allowing for a degree of intra-class variations~\cite{SANP11}.
However, as the closest points between two convex hulls are
artificially generated from linear combinations of certain
samples, deterioration in discrimination performance can occur if
the nearest points between two hulls are outliers or noisy. An
example is shown in Fig.~\ref{Fig:linear-generation}, where
unrealistic artificial variations are generated from combinations
of two distant samples.

Recent research suggest that creating multiple local linear models by clustering
can considerably improve recognition accuracy~\cite{Hadid09,Ruiping09,MMD08}.
In~\cite{Hadid04,Hadid09}, Locally Linear Embedding~\cite{LLE2000}
and $k$-means clustering are used to extract several representative exemplars.
Manifold Discriminant Analysis~\cite{Ruiping09}
and
Manifold-Manifold Distance~\cite{MMD08}
use the notion of maximal linear patches to extract local linear models.
For two sets with $m$ and $n$ local models,
the minimum distance between their local models determines the set-to-set distance,
which is acquired by $m \times n$ local model comparisons
(ie.~an exhaustive search).

A limitation of current multi-model approaches is that each set is clustered individually only once,
resulting in fixed clusters of each set being used for classification.
These clusters may not be optimal for discrimination and
may result in the two closest clusters representing two separate characteristics of an object.
For example, let us assume we have two face image sets of the same
person, representing two conditions. The clusters in the first set
represent various poses, while the clusters in the second set
represent varying illumination (where the illumination is
different to the illumination present in the first set). As the
two sets of clusters capture two specific variations, matching
image sets based on fixed cluster matching may result in a
non-frontal face (eg.~rotated or tilted) being compared against a
frontal face.

{\bf Contributions.} To address the above problems, we propose an
adaptive multi convex hull classification approach to find a
balance between single hull and nearest neighbour method.
The proposed approach integrates
affine/convex hull classification with an adapted multi-model
approach. We show that Maximum Margin Clustering (MMC) can be
applied to extract multiple local convex hulls that are distant
from each other. The optimal number of clusters is determined by
restricting the average minimal middle-point distance to control
the region of unrealistic artificial variations. The adaptive
reference clustering approach is proposed to enforce the clusters
of an gallery image set to have resemblance to the reference
clusters of the query image set.

\begin{figure}[!b]
  \vspace{-1ex}
  \centering
  \includegraphics[width=0.9\columnwidth]{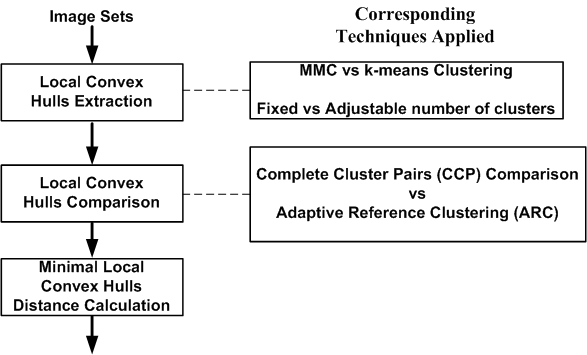}
  \caption
    {
    Framework of the proposed approach, including techniques used in conjunction with the proposed approach.
    }
  \label{Fig:flow-chart}
\end{figure}

 Consider two sets $S_a$ and $S_b$ to be compared.
The proposed approach first uses MMC to extract local convex hulls
from $S_a$ to diminish unrealistic artificial variations and to
constrain the noise in local convex hulls. We prove that after
local convex hulls extraction, the noisy set will be reduced. The
local convex hulls extracted from $S_a$ are treated as reference
clusters to constrain the clustering of $S_b$. Adaptive reference
clustering is proposed to force the clusters of $S_b$ to have
resemblance to the reference clusters in $S_a$,
by adaptively selecting the closest subset of images.
The distance of the closest cluster from $S_b$ to the reference cluster of $S_a$ is taken to indicate the distance between the two sets.
Fig.~\ref{Fig:flow-chart} shows a conceptual illustration of the proposed approach.
Comparisons on three benchmark datasets for face and object
classification show that the proposed method consistently
outperforms single hull approaches and several other recent
techniques. To our knowledge, this is the first method that
adaptively clusters an image set based on the reference clusters
from another image set.

We continue the paper as follows. In Section~\ref{sec:convex-hull}
and~\ref{sec:MMC}, we briefly overview affine and convex hull
classification and maximum margin clustering techniques. We then
describe the proposed approach in Section~\ref{sec:proposed},
followed by complexity analysis in Section~\ref{sec:complexity}
and empirical evaluations and comparisons with other methods in
Section~\ref{sec:experiments}. The conclusion and future research
directions are summarised in Section~\ref{sec:conclustion}.
\vspace{-1ex}
\section{Affine and Convex Hull Classification}
\label{sec:convex-hull}
\vspace{-1ex}
An image set can be represented with a convex model, either an affine hull or a convex hull, and then the similarity measure between two sets can be defined as the distance between two hulls~\cite{Hakan10}. This can be considered as an enhancement of nearest neighbour classifier with an attempt to reduce sensitivity of within-class variations by artificially generating samples within the set. Given an image set $S = [I_1, I_2, ... , I_n]$, where each $I_k, k\in [1,...,n]$ is a feature vector extracted from an image, the smallest affine subspace containing all the samples can be constructed as an affine hull model:

\noindent
\vspace{-1ex}
\begin{small}
\begin{equation}\label{eqn:affine-hull}
    H^{aff} = \{ y \}, \forall y = \sum\nolimits_{k=1}^{n}w_k I_k, \sum\nolimits_{k=1}^{n} w_k = 1.
\end{equation}%
\end{small}%

\noindent
Any affine combinations of the samples are included in this affine hull.

If the affine hull assumption is too lose to achieve good
discrimination, a tighter approximation can be achieved by setting
constraints on $w_k$ to construct a convex hull
via~\cite{Hakan10}:

\noindent
\vspace{-1ex}
\begin{small}
\begin{equation}\label{eqn:convex-hull}
    H^{con} \mbox{~=~} \{ y \}, \forall y \mbox{~=~} \hspace{-1ex} \sum\nolimits_{k=1}^{n} \hspace{-1ex}  w_k I_k, ~ \sum\nolimits_{k=1}^{n} \hspace{-1ex} w_k \mbox{~=~} 1, ~ w_k \in [0,1].
\end{equation}%
\end{small}

\noindent
The distance between two affine or convex hulls $H_{1}$ and $H_{2}$ are the smallest distance between any point in $H_1$ and any point in $H_2$:

\noindent
\begin{small}%
\begin{equation}\label{eqn:affine-dist}
    D_{h}(H_1,H_2) = \min_{y_1\in H_1, y_2 \in H_2} ||y_1 - y_2||.
\end{equation}%
\end{small}%

\noindent
In~\cite{SANP11}, sparsity constraints are embedded into the distance when matching two affine hulls to enforce that the nearest points can be sparsely approximated by the combination of a few sample images.
\vspace{-1ex}
\section{Maximum Margin Clustering}
\label{sec:MMC}
\vspace{-1ex}

Maximum Margin Clustering (MMC) extends Support Vector Machines (SVMs) from supervised learning to
the more challenging task of unsupervised learning. By formulating convex relaxations of the
training criterion, MMC simultaneously learns the maximum margin hyperplane and cluster
labels~\cite{MMC-05} by running an SVM implicitly. Experiments show that MMC generally outperforms
conventional clustering methods.

Given a point set {\small $S = [\Vec{x}_1, \Vec{x}_2, \ldots, \Vec{x}_n]$},
MMC attempts to find the optimal hyperplane and optimal labelling for two clusters simultaneously, as follows:

\noindent
\vspace{-2ex}
\begin{small}
\begin{eqnarray}\label{eqn:MMC}
    \min_{y \in [-1,+1]} \min_{\Vec{w},b,\xi_{i}} \frac{1}{2} \Vec{w}^{T}\Vec{w} + \frac{C}{n}\sum\nolimits_{i=1}^{n}\xi_{i} \hspace{8ex}\\
   s.t. \hspace{3mm} y_{i}(\Vec{w}^{T}\phi(\Vec{x}_{i})+b) \geq 1 - \xi_{i}, \hspace{1ex}   \xi_{i} \geq 0, ~~~ i = 1,\ldots,n. \nonumber
\end{eqnarray}%
\end{small}%

\noindent
where $y_i$ is the label learned for point $\Vec{x}_{i}$. Several approaches has been proposed to solve this challenging non-convex integer optimisation problem.
In~\cite{GMMC-07,MMC-05}, several relaxations and class balance constraints are made to convert the original MMC problem into a semi-definite programming problem to obtain the solution. Alternating optimisation techniques and cutting plane algorithms are applied on MMC individually in~\cite{Alternate-MMC-07} and~\cite{CPMMC-08} to speed up the learning for large scale datasets.
MMC can be further extended for multi-class clustering from multi-class SVM~\cite{multiclass-CPMMC-05}:

\vspace{-2ex}
\noindent
\begin{small}
\begin{eqnarray}\label{eqn:multi-class-MMC}
  \min_{y} \min_{\Vec{w}_{1},...,\Vec{w}_{k}, \xi} \frac{\beta}{2}\sum\nolimits_{p=1}^{k}||\Vec{w}_{p}||^{2}+\frac{1}{n}\sum\nolimits_{i=1}^{n}\xi_{i}\\ \nonumber
  s.t. \hspace{3mm} \Vec{w}_{y_{i}}^{T}\Vec{x}_{i} + \delta_{y_{i},r}-\Vec{w}_{r}^{T}\Vec{x}_{i} \geq 1 - \xi_{i} \\ \nonumber
  \forall i = 1,\ldots,n ~~~~~~ r = 1,\ldots,k.  \nonumber
\end{eqnarray}%
\end{small}%

\vspace{-1ex}
\section{Proposed Adaptive Multi Convex Hull}
\label{sec:proposed} \vspace{-1ex} The proposed adaptive multi
convex hull classification algorithm consists of two main steps:
extraction of local convex hulls and comparison of local convex
hulls, which are elucidated in the following text.
\vspace{-1ex}
\subsection{Local Convex Hulls Extraction (LCHE)}\label{subsec:LCHE}
\vspace{-1ex}
Existing single convex hull based methods assume that any convex combinations of samples represent intra-class variations and thus should also be included in the hull. 
However, as the nearest points between two hulls are normally
artificially generated from samples, they may be noisy or outliers
that lead to poor discrimination performance. On the contrary,
nearest neighbour method only compare samples in image sets
disregarding their combinations, resulting in sensitivity to
within class variations. There should be a balance between these
two approaches.

We observe that under the assumption of affine (or convex) hull
model, when sample points are distant from each other, linear
combinations of samples will generate non-realistic artificial
variations, as shown in Figure~\ref{Fig:linear-generation}. This
will result in deterioration of classification.

Given an image set with a noisy sample {\small $S = [\Vec{I}_1, \Vec{I}_2, \ldots, \Vec{I}_N, \Vec{I}_{ns}]$},
where $\Vec{I}_i, i\in 1, \ldots, N$ are normal sample images and {\small $\Vec{I}_{ns}$} is a noisy sample,
a single convex hull {\small $H$} can be constructed using all the samples.
The clear set {\small $H_{cl}$} of {\small $H$} is defined as a set of points whose synthesis does not require the noisy data {\small $\Vec{I}_{ns}$}.
That is,

\noindent
\begin{footnotesize}
\begin{equation}\label{eqn:clear-set}
    H_{cl}=\{p\}, \forall p = \sum\nolimits_{i} \alpha_{i} \Vec{I}_i, ~ \alpha_{i} \in [0,1], ~ \sum\nolimits_{i} \alpha_{i} = 1.
\end{equation}%
\end{footnotesize}%

\noindent
Accordingly, the remaining set of points in {\small $H$} not in {\small $H_{cl}$} is the noisy set {\small $H_{ns}$}.
The synthesis of points in noisy set must require the noisy data {\small $\Vec{I}_{ns}$}.
That is,
\noindent
\begin{footnotesize}
\begin{eqnarray}\label{eqn:noisy-set}
     H_{ns}=\{p_{ns} \}, \hspace{1ex} \forall p_{ns} = \alpha_{ns} \Vec{I}_{ns} +  \sum\nolimits_{i} \alpha_{i} \Vec{I}_i, \hspace{1ex} p_{ns} \notin H_{cl}, \\
     \alpha_{ns} \in (0,1], \hspace{1ex} \alpha_{i} \in [0,1], \hspace{1ex} \alpha_{ns} + \sum\nolimits_{i} \alpha_{i} = 1.\nonumber
\end{eqnarray}%
\end{footnotesize}%
\noindent
All of the normal samples $\Vec{I}_i$ that involve the synthesis of points in $H_{ns}$ are called the noisy sample neighbours, as they are generally in the neighbourhood of the noisy sample.
The noisy set $H_{ns}$ defines the set of points that is affected by the noisy data $\Vec{I}_{ns}$. If the nearest point $p$ between $H$ and other convex hulls lie in the noisy set, the convex hull distance is inevitably affected by $\Vec{I}_{ns}$.

\begin{figure}[b]
\vspace{-1ex}
  \centering

  \begin{minipage}{1.0\columnwidth}
    \begin{minipage}{0.45\textwidth}
      \centering
      \includegraphics[width=1\textwidth]{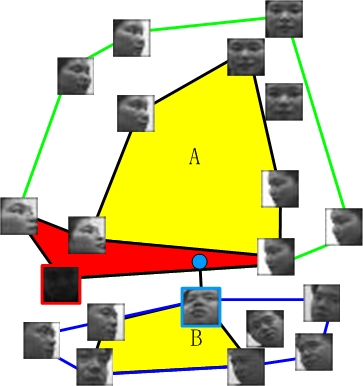}
    \end{minipage}
    \hfill
    \vline
    \hfill
    \begin{minipage}{0.45\textwidth}
      \centering
      \includegraphics[width=1\textwidth]{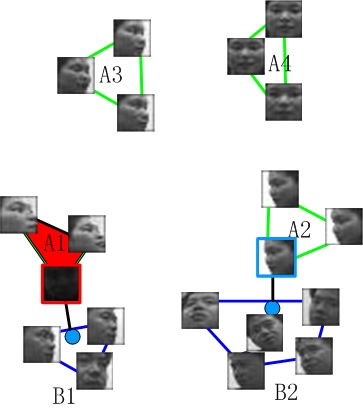}
    \end{minipage}
  \end{minipage}

  \begin{minipage}{1.0\columnwidth}
    ~
  \end{minipage}

  \begin{minipage}{1\columnwidth}
    \begin{minipage}{0.45\textwidth}
      \centering
      {\small\bf (a)}
    \end{minipage}
    \hfill
    \begin{minipage}{0.45\textwidth}
      \centering
      {\small\bf (b)}
    \end{minipage}
  \end{minipage}
\vspace{-1ex}
  \caption
    {
    Conceptual illustration of local convex hulls extraction. To compare image sets A and B,
    {\bf (a)}  existing single convex hull based methods use all samples in an image set to construct
    a model and calculate the distance between two closest points (indicated as the black line connecting the synthesised
    blue point and the real image with blue frame).
    The black image in set A is a noisy sample and the red area indicates the noisy set of A.
    As the blue point is synthesised from the combinations of the noisy sample and a real sample,
    it is also noisy. The yellow areas in A and B illustrate the region of unrealistic artificial
    variations generated from the combination of samples far away from each other.
    {\bf (b)} The proposed method divides image set A and B into 4 and 2 clusters separately and
    a local convex hull is constructed for each cluster individually.
    The black lines indicate the distance between image subsets A1 and B1, and A2 and B2.
    Notice that the closest points between subsets in (b) are completely different from those in (a).
    The blue points are artificially generated from real samples of subset B1 and B2 separately.
    As blue points are closely surrounded by the real sample points, they contain less artificial features.
    Moreover, after clustering, the noise set (red area) is significantly reduced and the
    unrealistic artificial variations (yellow area) are diminished.
    }
  \label{Fig:convex-hull-clustering}
\end{figure}

By dividing the single hull $H$ into multiple local convex hulls, {\small $H_1,H_2, \ldots, H_n$}, the noisy set is constrained to only one local convex hull. Assume the noisy sample is in one of the local convex hulls $H_{j}$, then the new noisy set $\widehat{H}_{ns}$ is defined as:
\noindent
\begin{footnotesize}
\begin{eqnarray}\label{eqn:new-noisy-set}
    \widehat{H}_{ns} \mbox{~=~} \{ p_{ns} \}, \hspace{1ex} \forall p_{ns}= \alpha_{ns} \Vec{I}_{ns} +  \sum\nolimits_{k} \alpha_{k} \Vec{I}_k, \hspace{1ex} p_{ns} \notin H_{cl}, \\
   \Vec{I}_{k} \in H_{j},  \hspace{1ex} \alpha_{ns} \in (0,1], \hspace{1ex} \alpha_{k} \in [0,1], \hspace{1ex} \alpha_{ns} + \sum\nolimits_{k} \alpha_{k} \mbox{~=~} 1. \nonumber
\end{eqnarray}%
\end{footnotesize}%
Comparing (\ref{eqn:new-noisy-set}) and~(\ref{eqn:noisy-set}), we notice that
\noindent
\begin{small}
\begin{equation}
    \widehat{H}_{ns} \subseteq H_{ns}.
\end{equation}%
\end{small}%
\noindent
Unless all the noisy sample neighbours are in the same local convex hull as the noisy sample, the noisy set will be reduced. By controlling the number of clusters to divide the noisy sample neighbours, the noise level can be controlled. Figure~\ref{Fig:convex-hull-clustering} illustrates the effect of local convex hulls extraction on noisy set reduction.

It is therefore necessary to divide samples from a set into
multiple subsets to extract multiple local convex hulls, such that
samples in each subset are similar with minimal artificial
features generated. Moreover, subsets should be far from each
other.
By dividing a single convex hull into multiple local convex hulls,
unrealistic artificial variations can be diminished and noise can
be constrained to local hulls.

A direct solution is to apply $k$-means clustering to extract local convex hulls~\cite{kmeans-2006}.
However, the local convex hulls extracted by $k$-means clustering are generally very close to each other,
without maximisation of distance between local convex hulls.
We propose to use MMC clustering to solve this problem.
For simplicity, we first investigate the problem for two local convex hulls.
Given an image set {\small $S = [\Vec{I}_1, \Vec{I}_2, \ldots, \Vec{I}_n]$},
images {\small $\Vec{I}_{i}, i \in 1,\ldots,n$} should be grouped into two clusters $C_{1}$ and $C_{2}$.
Two local convex hulls $H_{1}$ and $H_{2}$ can be constructed from images in the two clusters individually.
The two clusters should be maximally separated that any point inside the local convex hull is far from any point in the other convex hull.
It is equivalent to maximise the distance between convex hulls: 

\noindent
\vspace{-1ex}
\begin{small}
\begin{equation}\label{eqn:maximize-dist}
 \max_{C_1,C_2}D_{h}(H_{1},H_{2}).
\end{equation}%
\end{small}%

\noindent

The solution of Eqn.~(\ref{eqn:maximize-dist}) is equivalent to Eqn.~(\ref{eqn:MMC}).
Because finding the nearest points between two convex hulls is a dual problem
and has been proved to be equivalent to the SVM optimisation problem~\cite{Kristin-2000}.
Thus,

\noindent
\vspace{-1ex}
\begin{footnotesize}
\begin{eqnarray}\label{eqn:local-convex-hull}
\max_{C_1,C_2} D_{h}(H_{1},H_{2}) \mbox{~=~} \max_{C_1,C_2} \max_{\Vec{w},b, \gamma}  \gamma, \hspace{1ex} s.t. \hspace{1ex} \Vec{w}^{T}\Vec{I}_{k}+b \geq \gamma,   \\
                        \Vec{w}^{T}\Vec{I}_{m}+b \leq -\gamma, ~ ||\Vec{w}||_2 = 1, ~ \Vec{I}_{k} \in C_1, ~ \Vec{I}_{m} \in C_2. \nonumber
\end{eqnarray}%
\end{footnotesize}%
If we combine all of the images to make a set {\small $\Vec{I} = \{\Vec{I}_k, \hspace{1ex} \Vec{I}_m\}$},
then (\ref{eqn:local-convex-hull}) is equivalent to:
\noindent
\vspace{-1ex}
\begin{small}%
\begin{eqnarray}\label{eqn:local-convex-hull-reform}
\max_{C_1,C_2} D_{h}(H_{1},H_{2}) &  \mbox{~=~}  & \max_{y} \max_{\Vec{w},b, \gamma}  \gamma, \hspace{1ex} s.t. \hspace{1ex} y_i(\Vec{w}^{T} \Vec{I}_i + b) \geq \gamma,  \nonumber \\
                        &     & ||\Vec{w}||_2 = 1, y_i = \left\{^{+1, ~ \Vec{I}_i \in C_1}_{-1, ~\Vec{I}_i \in C_2} \right..
\end{eqnarray}%
\end{small}%
Maximisation of distance between clusters is the same as maximising the discrimination margin in Eqn.~(\ref{eqn:local-convex-hull-reform}) and is proved to be equivalent to Eqn.~(\ref{eqn:MMC}) in~\cite{MMC-05}.
We thus employ the maximum margin clustering method to cluster the sample images in a set to extract two distant local convex hulls.
Similarly, multi-class MMC can be used to extract multiple local convex hulls as in Eqn.~(\ref{eqn:multi-class-MMC}).

\vspace{-1ex}
\subsection{Local Convex Hulls Comparison (LCHC)}
\vspace{-1ex} In this section, we describe two approaches to
compare the local convex hulls: Complete Cluster Pairs Comparison
and Adaptive Reference Clustering.
\vspace{-1.5ex}
\subsubsection{Complete Cluster Pairs (CCP) Comparison}
\label{subsec:CCP}
\vspace{-1ex}
Similar to other multi-model approaches, local convex hulls extracted from two image sets can be
used for matching by complete cluster pairs (CCP) comparisons. Assuming multiple convex hulls
{\small $H_{a}^{1}, H_{a}^{2},..., H_{a}^{m}$} are extracted from image set {\small $S_{a}$},
and local convex hulls {\small $H_{b}^{1}, H_{b}^{2},..., H_{b}^{n}$} are extracted from image set {\small $S_{b}$}.
The distance between two sets {\small $S_{a}$} and {\small $S_{b}$}
is defined as the minimal distance between all possible local convex hull pairs:
\noindent
\vspace{-1ex}
\begin{small}
\begin{eqnarray}
    D_{ccp}(S_{a},S_{b}) \mbox{~=~} \min_{k,j} D_{h}(H_{a}^{k},H_{b}^{j}),  \forall k \in [1,m], \forall j \in [1,n].
\end{eqnarray}%
\end{small}%

Although LCHE can suppress noises in local convex hulls, CCP will
still inevitably match noisy data between image sets. Another
drawback of this approach is that fixed clusters are extracted
from each image set individually for classification. There is no
guarantee that the clusters from different sets capture similar
variations. Moreover, this complete comparison requires $m \times
n$ convex hull comparisons, which is computational expensive.

\vspace{-1.5ex}
\subsubsection{Adaptive Reference Clustering (ARC)}
\label{subsec:ARC}
\vspace{-1ex}

\begin{figure}[b]
\centering
  \begin{minipage}{1\columnwidth}
    \centering
       \begin{minipage}{1.0\textwidth}
          \begin{minipage}{0.45\textwidth}
            \centering{\footnotesize\bf set A}
          \end{minipage}
          \begin{minipage}{0.45\textwidth}
            \centerline{\footnotesize\bf set C}
          \end{minipage}
          \begin{minipage}{0.45\textwidth}
            \centerline{\includegraphics[width=\columnwidth]{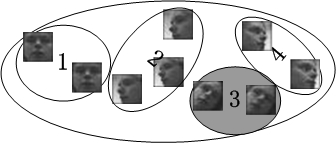}}
          \end{minipage}
          \begin{minipage}{0.05\textwidth}
            ~
          \end{minipage}
          \begin{minipage}{0.45\textwidth}
            \centerline{\includegraphics[width=\columnwidth]{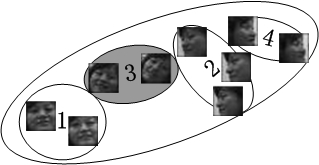}}
          \end{minipage}

       \end{minipage}
    \hfill
    \hrule
    \hfill
      \begin{minipage}{1\textwidth}
        ~
      \end{minipage}
    \begin{minipage}{1.0\textwidth}
      \begin{minipage}{0.45\textwidth}
        \centerline{\footnotesize\bf set B}
      \end{minipage}
      \begin{minipage}{0.45\textwidth}
        \centerline{\footnotesize\bf set C}
      \end{minipage}
      \begin{minipage}{0.45\textwidth}
        \centerline{\includegraphics[width=\columnwidth]{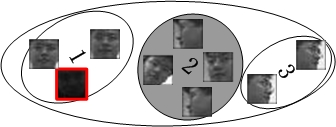}}
      \end{minipage}
      \begin{minipage}{0.05\textwidth}
        ~
      \end{minipage}
      \begin{minipage}{0.45\textwidth}
        \centerline{\includegraphics[width=\columnwidth]{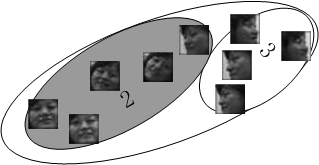}}
      \end{minipage}
    \end{minipage}

  \end{minipage}

  \caption
    {
    Conceptual illustration of adaptive reference clustering.
    Set~C is separately clustered according to the clusters of sets~A and~B.
    Set~C is divided into 4 clusters during comparison with set~A.
    The number in each cluster of set~C indicates that samples in this
    convex hull are close to the corresponding convex hull (with the same number) in set~A.
    The grey cluster pair indicates the two most similar clusters in sets~A and~C.
    Set~C is divided into 2 clusters when matching with set~B.
    Cluster 1 in set~B is a noisy cluster as it contains a noisy sample. All samples in
    set~C are far from the noisy cluster in set~B, therefore no samples are assigned to cluster 1.
    The light grey cluster pair indicates the closest two convex hulls between sets~B and~C.
    }
  \label{Fig:SANS-concept}
\end{figure}

In order to address the problems mentioned above, we propose the
adaptive reference clustering (ARC) to adaptively cluster each
gallery image set according to the reference clusters from the
probe image set (shown in Fig.~\ref{Fig:SANS-concept}). Assuming
image set {\small $S_{a}$} is clustered to extract local convex
hulls {\small $H_{a}^{1}, H_{a}^{2}, \ldots, H_{a}^{m}$}. For all
images {\small $I_{b}^{1}, I_{b}^{2}, \ldots, I_{b}^{N}$} from
image set {\small $S_{b}$}, we cluster these images according to
their distance to the reference local convex hulls from set
{\small $S_{a}$}.
That is, each image {\small $I_{b}^{i}$} is clustered to the closest reference local convex hull $k$: 

\noindent
\vspace{-2ex}
\begin{small}
\begin{equation}\label{eqn:reference-cluster}
    \min_{k} D_{h}(I_{b}^{i},H_{a}^{k})    \hspace{1ex} \forall k \in [1,m].
\end{equation}%
\end{small}%

After clustering all the images from set $S_{b}$, maximally $m$ clusters are obtained and local convex hulls
{\small $\widehat{H}_{b}^{1}, \ldots, \widehat{H}_{b}^{m}$} can then be constructed from these clusters.
Since each cluster {\small $\widehat{H}_{b}^{j}$} is clustered according to the corresponding reference cluster {\small $H_{a}^{j}$},
we only need to compare the corresponding cluster pairs instead of the complete comparisons. That is

\noindent
\vspace{-1ex}
\begin{small}
\begin{equation}
    D_{arc}(S_{a},S_{b}) = \min_{k} D_{h}(H_{a}^{k},\widehat{H}_{b}^{k}), \hspace{3mm} \forall k \in [1,m].
\end{equation}%
\end{small}%

ARC is helpful to remove noisy data matching (as shown in
Fig.~\ref{Fig:SANS-concept}). If there is a noisy cluster
$H_{a}^{n}$ in $S_{a}$, when no such noise exists in $S_{b}$, then
all the images $I_{b}^{i}$ are likely to be far from $H_{a}^{n}$.
Therefore, no images in $S_{b}$ is assigned for matching with the
noisy cluster.

\vspace{-1ex}
\subsection{Adjusting Number of Clusters}
\label{subsec:adjustable} \vspace{-1ex} One important problem for
local convex hull extraction is to determine the number of
clusters. This is because the \textit{convex combination} region
(\ie~regions of points generated from convex combination of sample
points) will be reduced as the number of clusters is increased (as
shown in Figure~\ref{Fig:convex-hull-clustering}). The reduction
may improve the system performance if the \textit{convex
combination} region contains many non-realistic artificial
variations. However, the reduction will negatively impact the
system performance when the regions are too small that some
reasonable combinations of sample points may be discarded as well.
An extreme case is that each sample point is considered as a local
convex hull, such as nearest neighbour method. We thus devise an
approach, denoted as average minimal middle-point distance (AMMD),
to indirectly ``measure'' the region of non-realistic artificial
variations included in the convex hull.

Let $I_a$ be a point generated from convex combination of two
sample points $I_i$, $I_j$ in the set $S$, $I_a = w_i I_i + w_j
I_j, w_i+w_j = 1, w_i, w_j > 0; I_i,I_j \in S$. The minimum
distance $\delta_i$ between $I_a$ to all the sample points in the
set $S$ indicates the probability of $I_a$ not being unrealistic
artificial variations. One extreme condition is when $\delta_i =
0$, then $I_a = I_k, I_k \in S$, which means $I_a$ is equivalent
to a real sample. The further the distance, the higher the
``artificialness'' of point $I_a$. In addition, the distance
between $I_a$ and $I_i$, $I_j$ must also be maximised in order to
avoid measurement bias. Thus, a setting of \mbox{$w_i= 0.5,
w_j=0.5$} (\ie~$I_a$ as the middle point) is applied. Having this
fact at our disposal, we are now ready to describe the AMMD.

For each sample point $I_i$ in the set $S$, we find $I_j$ which is
its furthest sample point in $S$. The minimal middle-point
distance of the sample point $I_i$ is defined via:
\begin{equation}\label{}
    \vspace{-1ex}
    \delta_i = \min_{k} ||I_a - I_k||_{2}, \forall I_k \in S.
\end{equation}
\noindent where $I_a$ is defined as $I_a = 0.5 I_i + 0.5 I_j$.
Finally, the AMMD of the set $S$ is computed via:
\begin{equation}\label{}
\vspace{-0.5ex}
    \Delta(S) = \frac{1}{N}\sum\nolimits_{i = 1}^{N} \delta_i.
\end{equation}
\noindent where $N$ is the number of sample points included in
$S$.

By setting a threshold $\Delta_{thd}$ to constrain AMMD, the
region of non-realistic artificial variations can be controlled.
An image set can be recursively divided into small clusters until
the AMMD of all clusters is less than the threshold.

\vspace{-1ex}
\section{Complexity Analysis}
\label{sec:complexity}
\vspace{-1ex}
Given two convex hulls with $m$ and $n$ vertexes, the basic Gilbert-Johnson-Keerthi (GJK) Distance
Algorithm finds the nearest points between them with complexity {\small $O(mn\log(mn))$}~\cite{GJK-97}.
The proposed approach needs a pre-processing step to cluster two image sets by MMC with a complexity of
{\small $O(sm)$} and {\small $O(sn)$}~\cite{multiclass-CPMMC-05}, where $s$ is the sparsity of the data.
Assuming each image set is clustered evenly with $M$ and $N$ clusters, to compare two local convex hulls,
the complexity is {\small $O(\frac{m}{M}\frac{n}{N}\log(\frac{m}{M}\frac{n}{N}))$}.
For the complete cluster pairs comparison, the total run time would be
{\small $O(MN\frac{m}{M}\frac{n}{N}\log(\frac{m}{M}\frac{n}{N})) + O(sm) + O(sn) = O(mn\log(\frac{mn}{MN})) + O(sm) + O(sn)$}.
The dominant term is on complete local convex hulls comparison, ie.~{\small $O(mn\log(\frac{mn}{MN}))$}.

By applying the adaptive reference clustering technique, one of the image set needs to be clustered
according to the reference clusters from the other set with a complexity of
{\small $O(nM\frac{m}{M}\log(\frac{m}{M}))$}.
Thus the total run time is
{\small $O(M\frac{m}{M}\frac{n}{N}\log(\frac{m}{M}\frac{n}{N})) +  O(sm) + O(nM\frac{m}{M}\log(\frac{m}{M})) = O(mn\log(\frac{(mn)^{1/N}}{(MN)^{1/N}}))+ O(sm) + O(mn\log(\frac{m}{M}))$}.
The dominant term is on the adaptive reference clustering, ie. {\small $O(mn\log(\frac{m}{M}))$}. \vspace{-1ex}
\section{Experiments}
\label{sec:experiments} \vspace{-1ex}
We first compare the design
choices offered by the proposed framework and then compare the
best variant of the framework with other state-of-the-art methods.

The framework has two components: Local Convex Hulls Extraction
(LCHE) and Local Convex Hulls Comparison (LCHC). There are two
sub-options for LCHE: (1) Maximum Margin Clustering (MMC) versus
$k$-means clustering (Section~\ref{sec:MMC}); (2) Fixed versus
Adjustable number of clusters (Section ~\ref{subsec:adjustable}).
There are also two sub-options for LCHC: (1) Adaptive Reference
Clustering (ARC) versus Complete Cluster Pairs (CCP)
(Section~\ref{subsec:ARC} and~\ref{subsec:CCP}); (2) Affine Hull
Distance (AHD) versus Convex Hull Distance (CHD)
(Section~\ref{sec:convex-hull}). We use two single hull methods as
baseline comparisons: the method using Affine Hull Distance
(AHD)~\cite{Hakan10} and the method using Convex Hull Distance
(CHD)~\cite{Hakan10}. Honda/UCSD~\cite{Honda-dataset} is used to
compare different variants of the framework to choose the best
one.

We use an implementation of the algorithm proposed
in~\cite{Yu-Feng09} to solve MMC optimisation problem. To
eliminate the bias on large number of images in one cluster, we
only select the top $m$ closest images to the reference cluster
for the ARC algorithm, wherein $m$ is the number of images in the
reference cluster. The clusters extracted from the query image set
are used as reference clusters to adaptively cluster each gallery
image set individually. In this way, the reference clusters stay
the same for each query, thus the distances between the query set
and each gallery set are comparable.

The best performing variant of the framework will then be
contrasted against the state-of-the-art approaches such as Sparse
Approximated Nearest Points (SANP)~\cite{SANP11} (a nearest point
based method), Mutual Subspace Method (MSM)~\cite{MSM-98} (a
subspace based method), and the Manifold Discriminant Analysis
(MDA) method~\cite{Ruiping09} (a multi-model based method). We
obtained the implementations of all methods from the original
authors. The evaluation is done in three challenging image-set
datasets: Honda/UCSD~\cite{Honda-dataset}, CMU
MoBO~\cite{CMU-MOBO-dataset} and the ETH-80~\cite{ETH-dataset}
datasets.

\vspace{-0.5ex}
\subsection{Datasets}
\vspace{-0.5ex}
We use the Honda/UCSD and CMU-MoBo datasets for face recognition tests. Honda dataset~\cite{Honda-dataset} consists of 59 video sequences of 20 subjects.
There are pose, illumination and expression variations across the sequences for each subject.
The CMU-MoBo dataset~\cite{CMU-MOBO-dataset} contains 96 motion sequences of 24 subjects with four walking patterns.
As in~\cite{MMD08}, face images from each frame of both face datasets were cropped and resized to \mbox{\small $20 \times 20$}.
We followed the protocol of~\cite{SANP11,Ruiping09} to conduct 10-fold cross validations on both datasets by randomly selecting one sequence for each
subject for training and using the rest for testing. On the Honda/UCSD dataset,
we tested on two types of image (raw and normalised via histogram equalisation),
using three configurations on the number of images: randomly chosen 50, randomly chosen 100, and all images.
Using a subset of images partly simulates real-world situations
where a face detector or tracker may fail on some frames.

The ETH-80 dataset~\cite{ETH-dataset} is used for object
recognition tests. It contains images of 8 object categories. Each
category includes 10 object subcategories (eg.~various dogs), with
each subcategory having 41 orientations. We resized the images to
\mbox{\small $32 \times 32$} and treated each subcategory as an
image set. For each category, we selected each subcategory in turn
for training and the remaining 9 for testing.

\vspace{-0.5ex}
\subsection{Comparative Evaluation}
\vspace{-0.5ex}

\begin{figure}[!b]
  \vspace{-2ex}
  \begin{minipage}{1\columnwidth}
    \begin{minipage}{0.49\textwidth}
      \centering
      {\it raw images}\\
      \vspace{0.5ex}
      \includegraphics[width=1\textwidth]{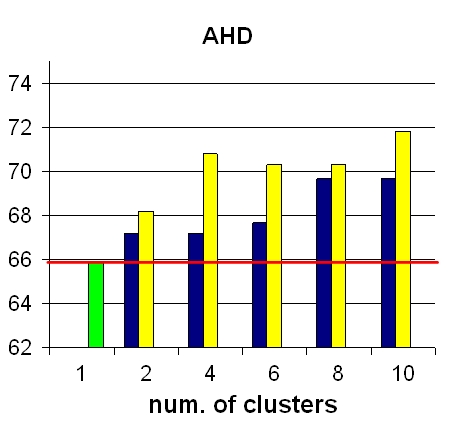}
      \includegraphics[width=1\textwidth]{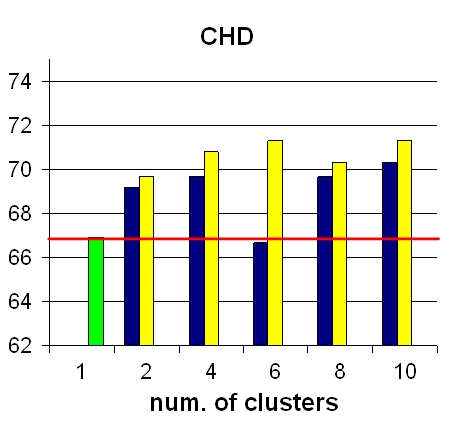}
    \end{minipage}
    %
    \vline
    \hfill
    \begin{minipage}{0.49\textwidth}
      \centering
      {\it normalised}\\
      \vspace{0.5ex}
      \includegraphics[width=1\textwidth]{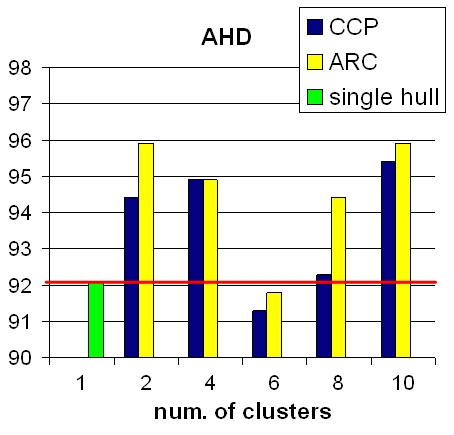}
      \includegraphics[width=1\textwidth]{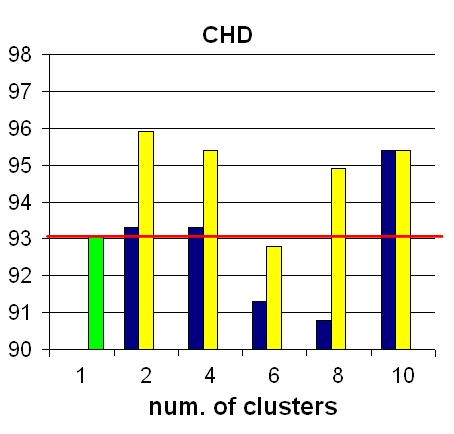}
    \end{minipage}
  \end{minipage}

  \caption
    {
        Performance comparison of the proposed local convex hulls extraction (LCHE) approach with single affine hull (AHD) and single convex hull (CHD) methods.
        Complete cluster pairs (CCP) and the adaptive reference clustering (ARC) techniques for local convex hulls comparisons are evaluated. Results are shown for the Honda/UCSD dataset with 100 images randomly selected per set. The left column contains results for raw images, while the right column for normalised images. The blue bar indicates the CCP technique and yellow bar is the ARC technique. The green bar is the result for the corresponding method using single hull, which can be considered as the baseline shown by the red line.
    }
  \label{fig:Honda-res-50}
\end{figure}

\begin{figure*}[!t]
  \centering
    \begin{minipage}{1\textwidth}
    \begin{minipage}{0.325\columnwidth}
      \centerline{\includegraphics[width=1\textwidth]{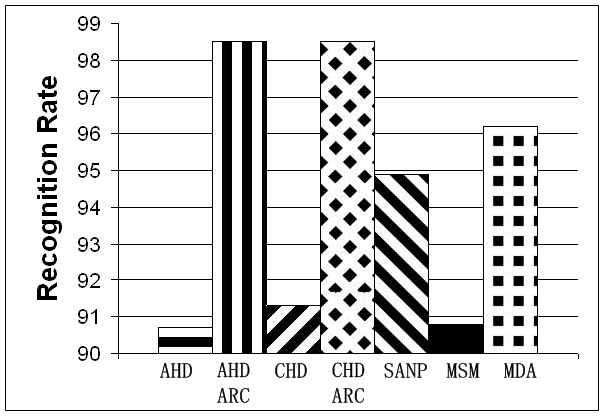}}
    \end{minipage}
    \hfill
    \begin{minipage}{0.325\textwidth}
      \centerline{\includegraphics[width=1\textwidth]{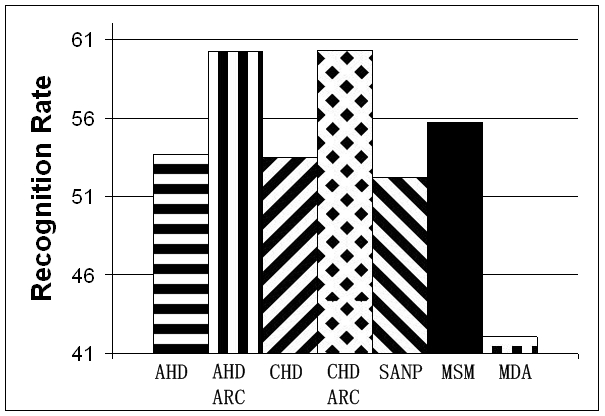}}
    \end{minipage}
    \hfill
    \begin{minipage}{0.325\textwidth}
      \centerline{\includegraphics[width=1\textwidth]{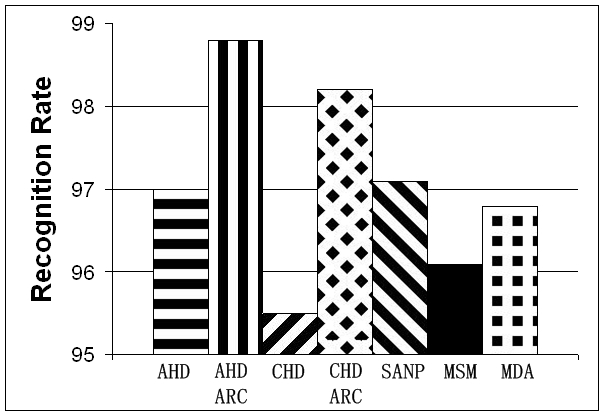}}
    \end{minipage}
  \end{minipage}

  \begin{minipage}{1\textwidth}
    \begin{minipage}{0.325\textwidth}
      \centerline{\bf \small Honda}
    \end{minipage}
    \hfill
    \begin{minipage}{0.325\textwidth}
      \centerline{\bf \small ETH-80}
    \end{minipage}
    \hfill
    \begin{minipage}{0.325\textwidth}
      \centerline{\bf \small CMU-MoBo}
    \end{minipage}
  \end{minipage}

  \caption
    {
        The proposed MMC and adjustable number of clusters combined with ARC technique compared with other methods,
        including SANP, MSM and MDA, on three datasets using all
        images. $\Delta_{thd}=5000$ is set for all datasets for the proposed method.
    }
  \label{fig:all-test-res}
  \vspace{-1ex}
\end{figure*}

We first evaluate the efficacy of the propose framework variants
using CCP and ARC. Here, the comparisons only choose MMC for
clustering using fixed number of clusters.
Figure~\ref{fig:Honda-res-50} illustrates the comparison of CCP
and ARC on AHD and CHD methods for the Honda/UCSD dataset with 100
images randomly selected per set.

The results show that under optimal number of clusters, ARC and
CCP outperforms the baseline counterparts (single hull with AHD
and CHD), indicating that LCHE improves the discrimination of the
system. ARC is consistently better than CCP regardless of which
number of cluster is chosen. This supports our argument that ARC
guarantees the meaningful comparison between local convex hulls
capturing similar variations.
As ARC performs the best in all tests, we thus only use ARC for
local convex hulls comparison in the following evaluations.

\begin{table}[!b]
  \vspace{-2ex}
  \hrule
  \vspace{0.5ex}
  \centering
  \caption
    {
    Comparing maximum margin clustering (MMC) vs \mbox{$k$-means}
    clustering as well as fix number vs adjustable number of clusters on local convex hulls extraction.
    The results are shown for AHD and CHD methods combined with adaptive reference clustering (ARC) technique on the Honda/UCSD dataset using 50 images
    per set. For fixed number of clusters, numbers shown in brackets indicate the corresponding optimal number of clusters.
    For adjustable number of clusters, numbers shown in square brackets are the optimal thresholds set for
    average minimal middle-point distance (AMMD).
    }
    \vspace{0.5ex}
  \label{tab:Honda-res-100}
      \begin{footnotesize}
              \begin{tabular}{c|c!{\vrule width 1.2pt}c|c!{\vrule width 1.2pt}c|c}
              \noalign{\hrule height 1.2pt}
               \multirow{2}{*}{cluster}   &  \multirow{2}{*}{cluster}   & \multicolumn{4}{c}{\bf image type}  \\
               \cline{3-6}
                     &         & \multicolumn{2}{c!{\vrule width 1.2pt}}{raw}  & \multicolumn{2}{c}{norm} \\
               \cline{3-6}
               \multirow{2}{*}{number}     & \multirow{2}{*}{methods}              &   AHD      &    CHD      &   AHD     &   AHD   \\
                                              &                       &   ARC      &    ARC      &   ARC     &   ARC   \\
               \noalign{\hrule height 1.2pt}
               \multicolumn{2}{c!{\vrule width 1.2pt}}{single hull}   &  68.4      &  69.5       &   94.6     &   92.8  \\
               \noalign{\hrule height 1.2pt}
               \multirow{4}{*}{fixed}   &  \multirow{2}{*}{MMC}       &  73.6      &  72.8       & 96.9           & 96.9 \\
                                        &                              &  (10)      &  (10)       &  (2)           & (2)   \\
                                        \cline{2-6}
                                        &  \multirow{2}{*}{$k$-means} &  71.5      &  69.7       &   96.9         & 96.4 \\
                                        &                              &  (2)       &  (4)        & (2)            & (2)  \\
               \noalign{\hrule height 1.2pt}
               \multirow{4}{*}{adjustable}&  \multirow{2}{*}{MMC}      &  {\bf 75.4}      & {\bf 74.9 }     & {\bf 99.5} & {\bf 97.9}  \\
                                          &                            &  [5000]          &  [5000]         & [30000]    & [30000]  \\
                                          \cline{2-6}
                                          &  \multirow{2}{*}{$k$-means}&  {71.8}          & {72.3 }         &  97.4      & 96.4   \\
                                          &                             & [5000]          & [5000]          & [30000]    & [30000]   \\
                \noalign{\hrule height 1.2pt}
              \end{tabular}
      \end{footnotesize}
\end{table}

The evaluation results of different variants for local convex
hulls extraction are shown in Table~\ref{tab:Honda-res-100}. We
only show the hyper-parameters which give the best performance for
each variant (\eg~the best number of cluster is shown in bracket
for fixed cluster number and the best threshold value is shown in
square bracket for adjustable cluster number). From this table, it
is clear that all the proposed variants outperform the baseline
single hull approaches, validating our argument that it is helpful
to use multiple local convex hulls. MMC variants outperform the
$k$-means counterparts indicating that maximising the local convex
hull distance for clustering leads to more discrimination ability
for system. The adjustable cluster number variant achieves
significant performance increase over fixed number of cluster
(between 1.5\% points to 5.2\% points). We also note that the
performance for adjustable number of clusters is not very
sensitive to the threshold. For instance, the performance only
drops by 2\% when the threshold $\Delta_{thd}$ is set to 5000 for
normalised images. In summary, MMC and adjustable number of
clusters combined with ARC achieve the best performance over all.

\begin{table}[!b]
    \vspace{-2ex}
    \hrule
    \vspace{0.5ex}
    \centering
    \caption
    {
        Comparing the proposed method with SANP, MSM, MDA and Nearest Neighbour (NN) on Honda data set with strong noise.
        80\% of the samples in each image set are replaced by noise.
    }
    \label{tab:noise}
    \vspace{1ex}
    \begin{footnotesize}
    \begin{tabular}{c|c|c|c|c|c|c|c}
        \noalign{\hrule height 1.2pt}
        \multirow{2}{*}{AHD}  &   AHD  & \multirow{2}{*}{CHD} & CHD  & \multirow{2}{*}{SANP}  & \multirow{2}{*}{MSM}  & \multirow{2}{*}{MDA} & \multirow{2}{*}{NN}\\
                              &   ARC  &                      & ARC  &                        &                       &                      &  \\
        \hline
        70.1        &   \bf{89.0}            &   77.9      & \bf{89.5}         & 77.2     &  80.0  & 76.2   & 82.1\\
        \noalign{\hrule height 1.2pt}
    \end{tabular}
    \end{footnotesize}
\end{table}

\begin{table}[!b]
    \centering
    \caption
    {
        Average time cost (in seconds) to compare two image sets on Honda dataset. The Convex Hull Distance (CHD) method combined with the proposed Complete Cluster Pairs (CCP) and Adaptive Reference Clustering (ARC) techniques are evaluated. `noc' indicates the number of clusters.
    }
    \label{tab:time}
    \vspace{1ex}
    \begin{small}
    \begin{tabular}{c|c|c|c|c|c}
        \noalign{\hrule height 1.2pt}
        {\bf num. of}     &   {\bf CHD}       &   \multicolumn{2}{c|}{\bf CHD CCP} &  \multicolumn{2}{c}{\bf CHD ARC}  \\
                         \cline{3-6}
        {\bf images}      &   \cite{Hakan10}  & noc = 2     & noc = 10     &   noc = 2    &  noc = 10  \\ \hline
        50          &   0.23            &   0.67      & 2.79         & 0.73     &  2.32  \\
        100         &   1.52            &  2.37       &  5.59        & 2.16     &  4.96  \\
        all         &   89.8            &  26.6       & 28.2        & 25.4    &  25.2    \\
        \noalign{\hrule height 1.2pt}
    \end{tabular}
    \end{small}
\end{table}

Results in Table~\ref{tab:noise} indicate that when strong noises
occur in image sets, the proposed ARC approach considerably
outperforms other methods, supporting our argument that ARC is
helpful to remove noisy data matching. It is worthy to note that
with strong noises, nearest neighbour (NN) method performs better
than single hull methods.

Figure~\ref{fig:all-test-res} is the summary of the best variant
found previously contrasted with the state-of-the-art methods.
Normalised images are used for Honda dataset and raw images are
used for ETH-80 and CMU-MoBo datasets. A fixed threshold
$\Delta_{thd}=5000$ is set for all datasets for adjustable number
of clustering. It is clear that the proposed system consistently
outperforms all other methods in all datasets regardless whether
AHD or CHD are used.

In the last evaluation, we compare the time complexity between the
variants. The average time cost to compare two image sets is shown
in Table~\ref{tab:time}. For small number of images per set,
extracting multiple local convex hulls is slower than using only
single convex hull because of extra time for MMC and adaptive
reference clustering. However, for large number (greater than 100)
of images per set, the proposed method is about three times faster
than the CHD method. That is because the number of images in each
cluster is significantly reduced, leading to considerably lower
time cost for local convex hulls comparisons.

\vspace{-1ex}
\section{Conclusions and Future Directions}\label{sec:conclustion}
\vspace{-1ex}
In this paper, we have proposed a novel approach to find a balance between single hull methods and nearest neighbour method.
 Maximum margin clustering (MMC) is employed to extract multiple local convex hulls for each query image set.
 The adjustable number of clusters is controlled by restraining the average minimal middle-point distance to constrain the
 region of unrealistic artificial variations. Adaptive reference clustering (ARC) is proposed to cluster the gallery image sets resembling the clusters of the query image set. Experiments on three datasets show considerable improvement over single hull methods as well as other state-of-the-art approaches. Moreover, the proposed approach is faster than single convex hull based method and is more suitable for large image set comparisons.

Currently, the proposed approach is only investigated for MMC and $k$-means clustering. Other clustering methods for local convex hulls extraction, such as spectrum clustering~\cite{Ncut-2000} and subspace clustering~\cite{Elhamifar09} and their effects on various data distributions need to be investigated as well.
\vspace{-1ex}
\begin{small}

~

\noindent
{\bf Acknowledgements.}
This research was funded by Sullivan Nicolaides Pathology, Australia and the Australian Research Council Linkage Projects Grant LP130100230.
NICTA is funded by the Australian Government through the Department of Communications and the Australian Research Council through the ICT Centre of Excellence program.
\end{small}

\bibliographystyle{ieee}
\noindent \vspace{-2.5ex}
\footnotesize
\bibliography{reference}

\end{document}